\documentclass[10pt,twocolumn,letterpaper]{article}

\usepackage{wacv}              

\usepackage{graphicx}
\usepackage{amsmath}
\usepackage{amssymb}
\usepackage{booktabs}

\usepackage[dvipsnames]{xcolor}
\usepackage{xspace}
\usepackage{multirow}
\usepackage{multicol}
\usepackage{soul}
\usepackage{pifont}
\usepackage{arydshln}

\newcommand\footnoteref[1]{\protected@xdef\@thefnmark{\ref{#1}}\@footnotemark}
\newcommand{\cmark}{\ding{51}}%
\newcommand{\xmark}{\ding{55}}%
\usepackage{float}

\newcommand{\todo}[1]{{\color{red}#1}}

\newcommand{\method}{TORE\xspace}

\newcommand{\iterator}{\textit{extractor}\xspace}
\newcommand{\Iterator}{\textit{Extractor}\xspace}
\newcommand{\aggregator}{\textit{aggregator}\xspace}
\newcommand{\midwtoks}{\textit{midway tokens}\xspace}
\newcommand{\Midwtoks}{\textit{Midway tokens}\xspace}

\usepackage{amsmath}
\usepackage{amssymb}

\newcommand{\X}{\mathcal{X}}
\newcommand{\Y}{\mathcal{Y}}

\newcommand{\Rep}{\mathcal{R}}

\newcommand{\vitb}{ViT--B\xspace}
\newcommand{\vitl}{ViT--L\xspace}

\newcommand{\clst}{\texttt{[CLS]}}

\DeclareMathOperator{\tore}{\method}

\DeclareMathOperator{\cache}{cache}

\usepackage[pagebackref,breaklinks,colorlinks]{hyperref}

\usepackage[capitalize]{cleveref}
\crefname{section}{Sec.}{Secs.}
\Crefname{section}{Section}{Sections}
\Crefname{table}{Table}{Tables}
\crefname{table}{Tab.}{Tabs.}

\def\wacvPaperID{815} 
\def\confName{WACV}
\def\confYear{2025}

\begin{document}

\title{TORE: Token Recycling in Vision Transformers for Efficient Active Visual Exploration}

\author{Jan Olszewski$^1$\\
\and
Dawid Rymarczyk$^{2,3}$\\
\and
Piotr Wójcik$^{4}$\\
\and
Mateusz Pach$^{2}$\\
\and
Bartosz Zieliński$^{2,5}$\\
\and
$^1$University of Warsaw\\
$^2$Faculty of Mathematics and Computer Science, Jagiellonian University
$^3$Ardigen SA \\ 
$^4$Center for Molecular Medicine Cologne (CMMC), University of Cologne
$^5$ IDEAS NCBR
}
\maketitle


\begin{abstract}
Active Visual Exploration (AVE) optimizes the utilization of robotic resources in real-world scenarios by sequentially selecting the most informative observations. However, modern methods require a high computational budget due to processing the same observations multiple times through the autoencoder transformers.
As a remedy, we introduce a novel approach to AVE called TOken REcycling (TORE). It divides the encoder into extractor and aggregator components. The extractor processes each observation separately, enabling the reuse of tokens passed to the aggregator. Moreover, to further reduce the computations, we decrease the decoder to only one block.
Through extensive experiments, we demonstrate that TORE outperforms state-of-the-art methods while reducing computational overhead by up to 90\%.
\end{abstract}

\section{Introduction}

In Active Visual Exploration (AVE)~\cite{pardyl2023active}, the agent selects consecutive observations based on a combination of learned policies and environmental cues to optimize information gain and task performance. It is essential in various applications assuming changing environments, such as rescue operations by drones~\cite{pirinen2023aerialviewlocalizationreinforcement}.

\begin{figure}
    \begin{center}
    \includegraphics[width=\linewidth]{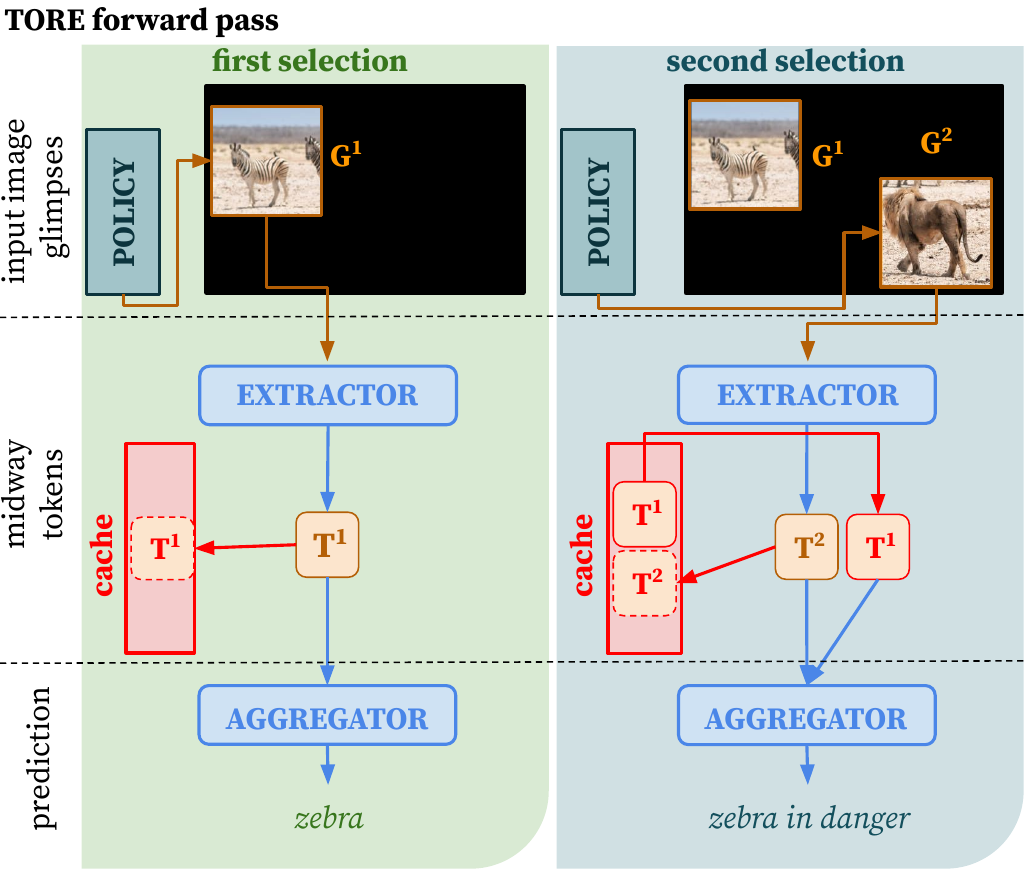}
    \end{center}
    \vspace{-1em}
    \caption{The forward pass of TORE for two sequential glimpses, denoted as $G^1$ and $G^2$. First, the glimpse $G^1$ is processed by the \iterator to generate the midway token $T^1$. This token is passed to the aggregator to obtain a prediction, but at the same time, it is cached for future use. When the glimpse $G^2$ appears, it is processed by the \iterator to generate the midway token $T^2$, which is passed to the aggregator together with the $T^1$ reused from the cache. This way, the \iterator processes each glimpse only once, significantly reducing the computations.}
    \label{fig:teaser}
    \vspace{-2em}
\end{figure}

Modern methods of AVE~\cite{pardyl2023active,huang2023glance} utilize Vision Transformers (ViT)~\cite{dosovitskiy2020vit}, which have profoundly reshaped computer vision by delivering efficiency on par with Convolutional Neural Networks (CNN)~\cite{li2022efficientformervisiontransformersmobilenet}. The key advantage of ViTs is their ability to handle incomplete inputs, in contrast to CNNs, which require data imputation before further analysis~\cite{przewikezlikowski2022misconv}. However, ViTs require a high computational budget~\cite{cai2022efficientvit}, which is problematic because AVE usually operates under strict resource constraints. Moreover, methods such as AME~\cite{pardyl2023active} and SimGlim~\cite{simglim} execute a full forward pass on all available observations, each time a new observation is collected.

One possible way to overcome this issue would be to replace ViTs in AVE with their efficient counterparts based on token pruning~\cite{dynamic_vit, avit} or architecture modifications~\cite{li2022efficientformervisiontransformersmobilenet, cai2022efficientvit}. However, they are not directly applicable to the AVE setup. Token pruning removes uninformative tokens but requires a global glance of the whole scene, which in AVE is yet to be discovered. Meanwhile, the architecture modifications limit the quadratic growth of self-attention computations but require structured and complete observations (full image, see Section 5.3 for more details), making the resulting model unsuitable for AVE purposes.

To address the computational inefficiency of the modern AVE methods while preserving their ability to process unstructured and incomplete observations, we introduce the TOken REcycling (\method) method. It divides ViT into two parts: \iterator and \aggregator (see Figure~\ref{fig:teaser}). The \iterator processes glimpses separately, generating \midwtoks that are cached. The \aggregator combines the \midwtoks and returns a prediction. This way, the results of computations made by the \iterator are reused, substantially reducing the computational burden compared to previous AVE methods.
%
Additionally, we advocate for using a lightweight decoder since perfect image reconstruction is unnecessary to achieve satisfactory classification.

By conducting comprehensive experiments, we demonstrate that \method{} significantly cuts computational overhead by up to 90\% with improved accuracy compared to the state-of-the-art methods for AVE. Furthermore, employing thorough ablations and analysis, we motivate our architectural decisions and show the trade-off between efficiency and precision. The code is available at \url{https://github.com/jano1906/TokenRecycling}.

Our contributions can be summarized as follows:
\begin{itemize}
    \item We introduce TORE, an efficient method for Active Visual Exploration, which substantially reduces the computational burden.
    \item We propose a training strategy with a random sampling of glimpses, increasing the model's generalizability.
    \item We conduct extensive experiments on AVE benchmarks showing the accuracy-efficiency trade-off.
\end{itemize}

\section{Related Works}
\paragraph{Active Visual Exploration.} 
AVE involves an agent guiding its view to successively improve internal representations of the observed world~\cite{activevision,exploring_unseen}, what can be seen in the context of simultaneous localization and mapping (SLAM)~\cite{ramakrishnan2021exploration}. In this context, the selection strategies for successive observations (exploration) play a crucial role. Multiple approaches are proposed to address this challenge, such as image reconstruction techniques~\cite{exploring_unseen,bayesian,360degrees}, dedicated exploration models~\cite{rangrej2021visual,rangrej2022consistency,simglim}, the utilization of self-attention mechanisms~\cite{seifi2021glimpse}, and the analysis of entropy~\cite{pardyl2023active}. AVE has also been explored in the context of segmentation~\cite{attend_and_segment} and 3D vision~\cite{jayaraman2018learning}.

Our work contributes to this landscape by introducing a novel approach dedicated to AVE that recycles computations within a transformer-based architecture. 

\paragraph{Efficient computations for vision models.} Recent advancements in computer vision show that the increase in model accuracy is often related to the increase in its computational needs. That is why a lot of research is done to reduce the amount of computations and at the same time preserve the high fidelity of the model. One research branch is focused on efficient training techniques, including continual learning~\cite{masana2022class,rymarczyk2023icicle}, domain adaptation~\cite{varsavsky2020test,wang2022continual}, and fine-tuning~\cite{he2023sensitivity,liang2022expediting}. Another one on reducing the inference computations of models, such as early exits~\cite{passalis2020efficient,wolczyk2021zero}, dedicated to a given problem models, e.g., in medical imaging~\cite{raghu2019transfusion}, and efficient architectures~\cite{sandler2018mobilenetv2,tan2019efficientnet}.

In the scope of this work, we are particularly interested in methods assuring efficient ViT inference. Approaches aiming at the reduction of ViT's computational needs propose token pruning~\cite{dynamic_vit,Yin_2022_CVPR,avit}, low-rank factorization~\cite{compressing}, limiting self-attention to non-overlapping local windows~\cite{liu2021Swin}, scaling attention mechanism with sequence length~\cite{Beltagy2020Longformer}, grouping attention layers into local and global information exchange layes~\cite{cai2022efficientvit, li2022efficientformervisiontransformersmobilenet}, and replacing fully-connected layers with a star-shaped topology~\cite{star_transformer}.

Our method (\method) aligns with recent trends aimed at enhancing the computational efficiency of Vision Transformers. Specifically, we modify the ViTs' forward pass to optimize for Active Visual Exploration because current efficient architectures are not well-suited for processing incomplete and unstructured inputs found in the AVE task.
 
\section{Method}

\begin{figure*}[t]
    \centering
    \includegraphics[width=0.95\textwidth]{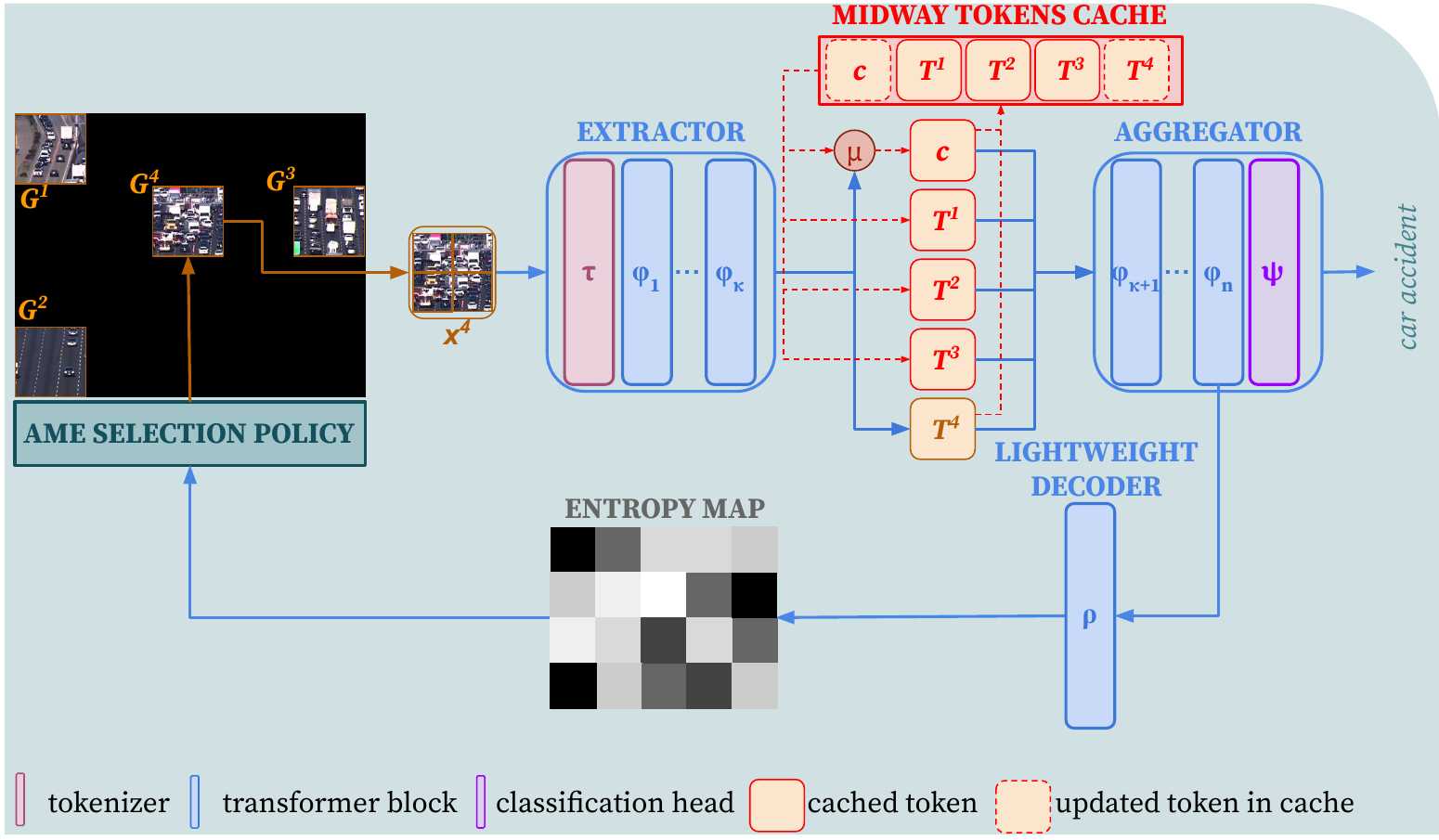}
    \caption{\method is an efficient approach to Active Visual Exploration based on vision transformers. In each step, it uses \textit{AME selection policy} to collect the next glimpse (e.g. $G^4$) based on the entropy map generated based on previous glimpses ($G^1,\dots,G^3$). Then, $G^4$ is divided into patches, which are tokenized and pushed through the \iterator. \Midwtoks are stored in the cache and passed to the \aggregator to obtain a prediction. Thanks to the cache of tokens, each glimpse is processed by \iterator only once, which reduces the computational overhead. Please note that the classification token $c$ is defined as the average from all classification tokens generated by the \iterator.}
    \label{fig:method}
    \vspace{-2em}
\end{figure*}

\method aims to reduce the computational budget in the AVE solution based on Vision Transformers (ViT) by reusing tokens. Therefore, to make this paper self-contained, we start this section by recalling ViT and AVE. After that, we describe our method and its training policy.

\paragraph{Vision Transformers.} Transformer model $M$ of depth $n$ consists of:
\begin{itemize}
    \item tokenizer $\tau: \X \to \Rep$,
    \item transformer blocks $\varphi_i : \Rep \to \Rep$, for $i\in\{1, \dots, n\}$,
    \item prediction head $\psi: \Rep \to \Y$.
\end{itemize}
where $\X$ is an input domain (i.e. set of images), $\Y$ is an output domain (i.e. logits), and $\Rep$ is a representation space. For a given $x \in \X$, tokenizer $\tau$ generates a token set that contains $\clst$, a special global information token denoted as $c$, and $T$, a set of remaining patch-corresponding tokens.

\paragraph{AVE sequential prediction.}
Suppose that input $x$ consists of $p$ glimpses which we receive sequentially, ie. $x=\{x^{(1)}, \dots, x^{(p)}\}$ and at time $j$, the glimpses $x^{(i)}$ for $i\leq j$ are available. In such case, we are interested in predictions $M^{(j)}(x)~=~M\left(x^{(1)},\dots, x^{(j)}\right)$ recomputed each time when a new glimpse $x^{(j)}$ is selected, for $1\leq j\leq p$. The consecutive glimpses are selected with a \textit{selection policy}.

\subsection{TOken REcycling (TORE)}

\paragraph{\Iterator-\aggregator framework.}
As presented in Figure~\ref{fig:method}, we divide the transformer $M$ into two paths \iterator $M_E$ and \aggregator $M_A$:
\begin{align}
    M_E &= \varphi_{\kappa} \circ \dots \circ \varphi_1 \circ \tau \nonumber \\
    M_A &= \psi \circ \varphi_n \circ \dots \circ \varphi_{\kappa + 1}, 
\end{align}
where a parameter $\kappa \in \{0,1, \dots, n\}$ is the number of transformer blocks assigned to \iterator $M_E$, while \aggregator $M_A$ is built from remaining $n-\kappa$ blocks.
Given a sample $x^{(j)}$, the \iterator produces \midwtoks, which we denote as $\{c^{(j)}\}\cup T^{(j)} := M_E(x^{(j)})$. We define the TORE Forward Pass (TORE FP) with parameter $\kappa$ at time $j$ as 
\begin{equation}
\small
\tore_{\kappa}^{(j)}(x) = M_A \left( \Lambda\left\{ M_E(x^{(1)}),\dots, M_E(x^{(j)})\right\}\right),
\label{eq:tore}
\end{equation}
where 
\begin{equation}
\small
\Lambda\left\{ M_E(x^{(1)}),\dots, M_E(x^{(j)})\right\} = \Bigl\{\frac{1}{j}\sum_{i=1}^j c^{(i)}\Bigr\} \cup \bigcup_{i=1}^j T^{(i)}.   
\end{equation}
Therefore, we apply the \iterator to each input glimpse $x^{(i)}$. Then, we combine the \midwtoks by averaging $c^{(i)}$ tokens and taking the union over glimpse-corresponding tokens $T^{(i)}$. The final output is produced with the \aggregator.

\paragraph{Efficient sequential inference with TORE forward pass.}
To efficiently compute the Eq.~\ref{eq:tore}, we cache outcomes of already calculated forward passes at times $i<j$. We update the cache as follows:
\begin{align}
    \cache^{(0)}_c &= 0 \nonumber \\
    \cache^{(0)}_T &= \emptyset \nonumber\\
    \cache^{(j)}_c &= \frac{1}{j}c^{(j)} + \frac{j-1}{j}\cache^{(j-1)}_c \nonumber\\
    \cache^{(j)}_T &= T^{(j)} \cup \cache^{(j-1)}_T,
\end{align}
and then calculate the prediction as:
\begin{equation}
\tore_\kappa^{(j)}(x) = M_A(\cache^{(j)}).
\end{equation}

This way, the consecutive values of $\tore_{\kappa}^{(j)}(x)$ are computed by processing each $x^{(j)}$ with the \iterator only once. It is in contrast to ViT's original forward pass where at each timestep $j$ a full collection of inputs $\{x^{(1)}, \dots, x^{(j)}\}$ is processed by the network $M$.

\begin{figure}[t]
   \centering
   \includegraphics[width=\linewidth]{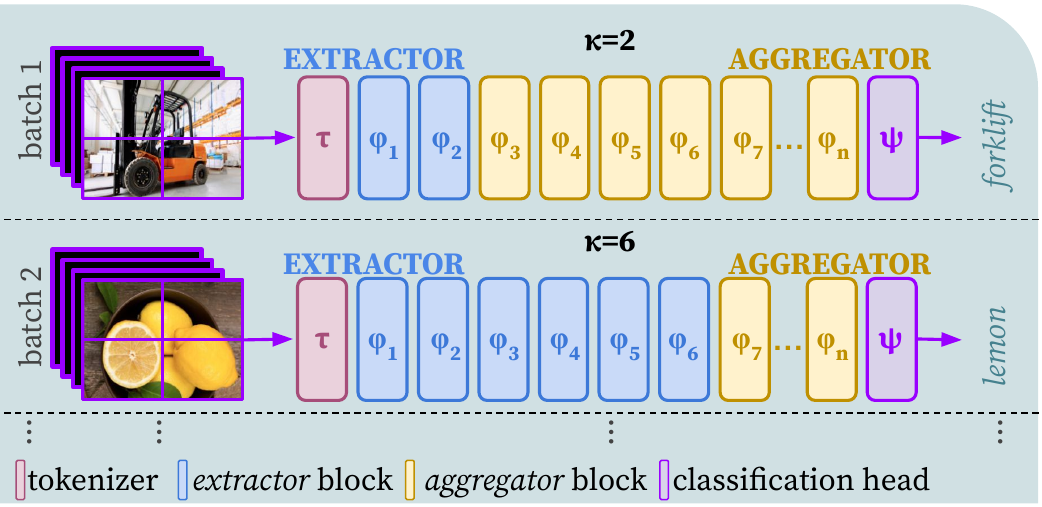}
   \caption{Our training policy, in which we sample $\kappa$ responsible for the size of \iterator and \aggregator from the uniform distribution for each batch to ensure the model's flexibility during inference. In this example, the number of \iterator blocks equals 2 and 6 for the top and bottom parts, respectively.}
   \label{fig:training}
  \vspace{-1em}
\end{figure}

\vspace{-1em}

\paragraph{Attention Map Entropy (AME).}
We adopt the AME~\cite{pardyl2023active} policy to guide the glimpse selection in sequential prediction during inference. To pick glimpse $x^{(j + 1)}$ at the time $j + 1$, the policy utilizes an additional transformer-based reconstruction decoder $\rho$. It is fed with embeddings of previously chosen glimpses $x^{(1)}, \ldots, x^{(j)}$, and mask embeddings corresponding to the missing glimpses as in original MAE~\cite{MaskedAutoencoders2021} work. The embeddings are obtained from the last block of the \aggregator, before applying the prediction head $\psi$.

Besides the reconstructed image, the decoder also provides attention map $A^{(j)}_{h}$ from each head $h$ in its last transformer block, used to calculate the entropy map $H^{(j)}$ with the following formula:
\begin{align}
    H^{(j)} &= \sum_h H(A^{(j)}_{h}),
\end{align}
where $H(A) = \{-\sum_{j} A_{ij}\log_2(A_{ij})\}_{i}$ is a row-wise Shannon entropy calculated for attention matrix $A$.

After zeroing entries of $H^{(j)}$ corresponding to the already chosen glimpses, AME selects the new glimpse that has the highest value in $H^{(j)}$. Intuitively, the policy picks the glimpse with the highest reconstruction uncertainty, assuming it is the most informative for the prediction.

\vspace{-1em}
\paragraph{Lightweight decoder.}
Contrary to the AME~\cite{pardyl2023active}, our reconstruction decoder used to calculate the entropy map is tiny. It has only 1 transformer block with 4 heads and 128 hidden dimensions, instead of 8 blocks with 16 heads and 512 hidden dimensions used originally~\cite{MaskedAutoencoders2021, pardyl2023active}.
\vspace{-1em}
\paragraph{Training with Random Glimpse Selection Policy.}
Our training sample $(x=\{x^{(1)}, \ldots, x^{(q)}\},w,y)$ comprises $q$ randomly picked image glimpses, a whole image, and a label. It is in contrast to AME~\cite{pardyl2023active}, which uses the entropy map to select the glimpses $x$ at the training time as well as at the inference time. The number of glimpses $q$ is fixed for all samples and set to occupy $\sim25\%$ of the whole image.
\vspace{-1em}
\paragraph{Random $\kappa$ sampling.}
For each training batch of $B$ such triplets, we sample the size of the \iterator from the uniform distribution $\kappa~\sim~\mathcal{U}\{0,\dots, n\}$, as illustrated in Figure~\ref{fig:training}. Next, we perform a forward pass of $\tore_\kappa^{(q)}(x_i)$ to get class prediction $\hat{y}_i$ and reconstructed image $\hat{w}_i$, for $1\leq i \leq B$. Finally, we compute loss function $L$ combining cross-entropy classification loss $L_c$ and root-mean-squared error reconstruction loss $L_r$:
\begin{align}
     L_c &= - \sum_{i=1}^B y_i \cdot \log(\hat{y}_i) \nonumber\\
     L_r &= \sum_{i=1}^B \sqrt{\frac{1}{P} \sum_{j=1}^{P} (p(w_i, j) - p(\hat{w}_i, j))^2} \nonumber\\
     L &= L_c + \lambda L_r,
\end{align}
where $\lambda=0.1$ is a weighting factor, $p(w, j)$ denotes $j$-th pixel of image $w$ and $P$ is the total number of pixels.

In consequence, we obtain a single model that reliably works for multiple values of $\kappa$ during inference, making the method flexible. This property allows us to balance accuracy and efficiency on demand, i.e., by increasing the \iterator size $\kappa$, we can reduce computations but decrease the model's accuracy.

\section{Experimental Setup}

\label{sec:exp_set}

\begin{figure*}[t]
    \centering
    \includegraphics[width=\textwidth]{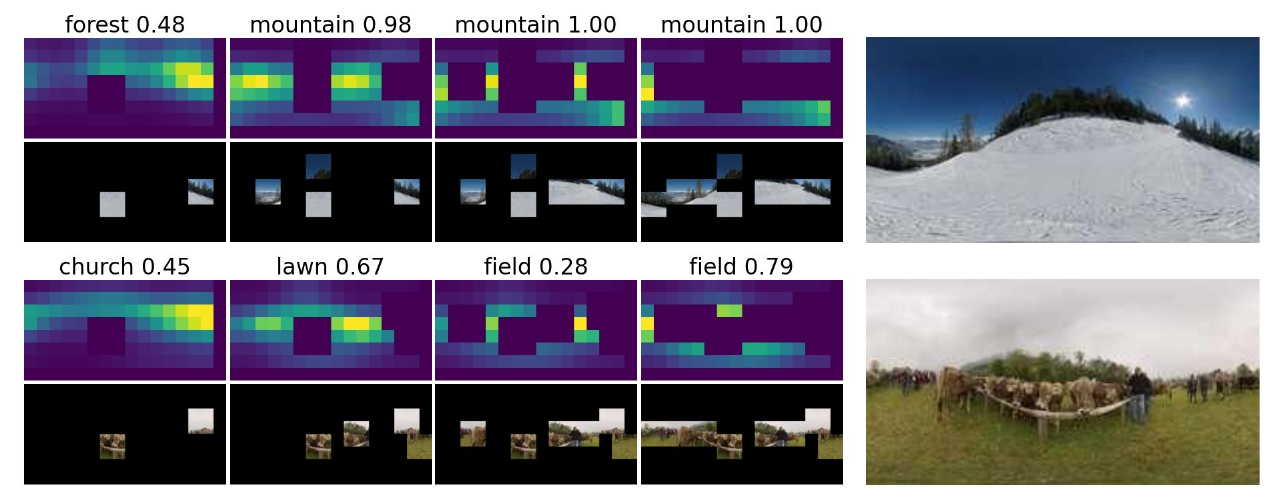}
    \caption{Visualization of active visual exploration performed by TORE. On the left side, we present entropy maps used by the selection policy, its successive fields of view, and the corresponding predictions. On the right side, we present the original images.}
    \label{fig:prediction_visualization}
    \vspace{-1em}
\end{figure*}

\subsection{Implementation Details}
We implement the \iterator-\aggregator framework using the transformer $M$ based on the \vitb architecture consisting of $n=12$ blocks. The model is initialized with MAE~\cite{MaskedAutoencoders2021} weights pretrained on ImageNet-1k~\cite{deng2009imagenet}. The originally learned positional encodings are replaced with sinusoidal ones, following implementation introduced in AME~\cite{pardyl2023active}. Our lightweight decoder is a downsized modification of the MAE decoder with randomly initialized weights. Both networks use $16 \times 16$ pixel-sized tokens.

All models are trained for a maximum of $100$ epochs, stopping if validation loss does not improve for consecutive $10$ epochs. We select weights achieving the lowest validation loss. The batch size $B$ equals $16$. The optimizer of our choice is AdamW~\cite{loshchilov2017decoupled} with beta values of $0.9$ and $0.999$, epsilon of $10^{-8}$, and weight decay of $0$. The learning rate, dynamically adjusted via the cosine scheduler with a minimum value of $10^{-8}$, is initialized at $10^{-5}$. We use the same data augmentations as in AME~\cite{pardyl2023active}, namely the $Random HorizontalFlip$ and $RandomResizedCrop$ with scale in range $(0.2, 1.0)$.

\subsection{Active Visual Exploration}
\label{sec:exp_ave}
We test our approach on SUN360~\cite{xiao2012recognizing}, CIFAR100~\cite{krizhevsky2009learning}, Food101~\cite{kaur2017combiningweaklyweblysupervised} and Flowers102~\cite{4756141}  datasets. SUN360 is a scene recognition dataset with 26 classes, containing both indoor and outdoor images captured with 360$^{\circ}$ camera, making a suitable evaluation dataset for vision algorithms dedicated to robotics applications. The other considered datasets are object-centric image classification datasets with 100, 101, and 102 classes respectively. We keep native $128\times 256$ resolution for the SUN360 dataset for a fair comparison to the baselines and resize images from the other datasets to $224\times 224$ resolution. As the SUN360 dataset does not provide a predefined train-test split, we divide the dataset into train-test with a ratio of 9:1 based on image index, following the methodology used in \cite{pardyl2023active, seifi2021glimpse}. For the other datasets, we use the original train-test split.

During training and evaluation, we sample non-overlapping $2\times 2$ token-sized glimpses, which in total cover about $25\%$ of an input image (8 glimpses for SUN360 and 12 glimpses for the other datasets). When training, we sample the glimpses uniformly at random. During evaluation, we sample the glimpses according to the chosen policy.

We follow the evaluation protocol introduced in~\cite{pardyl2023active} for all experiments in the AVE setup.

\begin{table*}[t]
    \centering
    \begin{tabular}{cccccc}
        \toprule
        Dataset & Method & Architecture & Glimpse Regime  & GFLOPs $\downarrow$ & Accuracy $\uparrow$ \\ \midrule
        \multirow{6}{*}{SUN360} &AttSeg~\cite{attend_and_segment} & custom & \multirow{2}{*}{$8\times 48^2$} & NA\footnote{Due to the lack of shared code and details in the AttSeg paper, we are not able to estimate the required GFLOPs.} & 52.6 \\
        &GlAtEx~\cite{seifi2021glimpse} & ResNet-18 &  & 55.8 & 67.2 \\ \cline{2-6}
        &AME~\cite{pardyl2023active} & \vitl & \multirow{4}{*}{$8\times 32^2$} &  71.9 &73.4 \\
        &AME & \vitb & &  38.8 & 64.3 \\
        &TORE$_{0}$ & \vitb & & 13.4 & \textbf{77.1} \\
        &TORE$_{8}$ & \vitb &  & \textbf{7.0} & 74.1 \\ \midrule
        \multirow{3}{*}{CIFAR100} &AME  & \multirow{3}{*}{\vitb}  & \multirow{3}{*}{$12\times 32^2$} & 87.0 & 73.0 \\ 
        & TORE$_0$ & &  & 29.1 & \textbf{80.2} \\
        & TORE$_7$ & &  & \textbf{15.9} & 73.5 \\ \midrule
        \multirow{3}{*}{FLOWERS102} &AME  & \multirow{3}{*}{\vitb} & \multirow{3}{*}{$12\times 32^2$} & 87.0 & 72.4 \\
        & TORE$_0$ & &  & 29.1 & \textbf{86.3} \\
        & TORE$_6$ & &  & \textbf{16.4} & 84.4 \\ \midrule
        \multirow{3}{*}{FOOD101} &AME  & \multirow{3}{*}{\vitb} & \multirow{3}{*}{$12\times 32^2$} & 87.0 & 74.2 \\
        & TORE$_0$ & &  & 29.1 & \textbf{80.4} \\
        & TORE$_6$ &  &  & \textbf{16.4} & 78.4 \\
        \bottomrule
    \end{tabular}
    \caption{Comparison of the accuracy and computational efficiency across various AVE methods reveals the superiority of our TORE over the baseline methods. In fact, TORE achieves higher accuracy than AME even with a much smaller ViT-B architecture (compared to ViT-L in AME) with significantly reduced computational load. Furthermore, by adjusting the value of $\kappa$ up to 8 or 6, we can reduce computations by an additional 50\% while still surpassing other methods' performance.}

    \label{tab:main_table}
\end{table*}

\begin{table*}
    \centering
    \begin{tabular}{ccccccc}
        \toprule Method & Architecture & Glimpse Regime & Image Res. & GFLOPs $\downarrow$ & Reconstruction $\downarrow$ \\ \midrule
        AttSeg & custom & \multirow{3}{*}{$8\times 48^2$} & \multirow{3}{*}{$128\times 256$} & NA\todo{\footnotemark[\value{footnote}]} &37.6 \\
        GlAtEx & ResNet-18 &  &  & 55.8 & 33.8 \\
        AME & ViT-L &  &  & 101.1 &\textbf{23.6} \\ \midrule
        SimGlim\cite{simglim} & NA\todo{\footnotemark[\value{footnote}]} & \multirow{2}{*}{$37\times 16^2$} & \multirow{2}{*}{$224\times 224$} & NA\todo{\footnotemark[\value{footnote}]} & 26.2 \\
        AME & ViT-L &  & & 390.6 & \textbf{23.4} \\ \midrule
        AME & ViT-L & \multirow{3}{*}{$8\times 32^2$} & \multirow{3}{*}{$128\times 256$} & 71.9& \textbf{29.8} \\
        TORE$_{0}$ & ViT-B &  & & 13.4 & 40.6 \\
        TORE$_{8}$ & ViT-B &  & & \textbf{7.0} & 43.2 \\
        \bottomrule
        \end{tabular}
        \caption{Reconstruction results in the AVE task for SUN360. Due to focusing TORE on efficient classification and using a lightweight decoder, the reconstruction performance is relatively poor. However, at the same time, the computational burden is significantly reduced. }
    
    \label{tab:reconstructions}
    \vspace{-1em}
    \end{table*}

\section{Results}
In this section, we present the results for two tasks within Active Visual Exploration: image classification and reconstruction. Subsequently, we conduct thorough ablations and analysis of our TORE method. 

Before delving into the metrics for these tasks, we illustrate how AVE is performed at selected timesteps, as depicted in Figure~\ref{fig:prediction_visualization}. It can be observed that the entropy map of the reconstruction decoder guides the model in a way tailored to each sample. As the model observes a larger portion of relevant input, it is more certain of a given prediction.

\subsection{Classification}
TORE exhibits superior performance in the classification task compared to other AVE methods, as demonstrated in Table~\ref{tab:main_table}. Our approach achieves notable reductions in computations of up to 90\% for SUN360$_{128\times 256}$ and 80\% for the other datasets while maintaining state-of-the-art accuracy. Particularly noteworthy is that our method with a ViT-B backbone outperforms AME utilizing a ViT-L backbone. To ensure a fair comparison, we present the accuracy of AME employing a ViT-B encoder and its original decoder size.

The incorporation of a lightweight decoder and a random glimpse selection policy during training significantly enhances the classification task's performance while concurrently reducing the exploration's computational costs. Additionally, the \iterator-\aggregator framework for the ViT backbone in TORE provides flexibility through various $\kappa$ values, allowing prioritization of either higher accuracy or improved resource utilization.

\subsection{Reconstruction}

As our method focuses on the classification task, the reconstruction can be seen solely as an auxiliary task. In Table~\ref{tab:reconstructions}, we present the reconstruction RMSE of our method compared to other state-of-the-art approaches. Interestingly, the performance of a model on the reconstruction task within the AVE setup does not necessarily correlate with its classification performance. Despite the significant reduction in reconstruction capabilities due to the use of a lightweight decoder in TORE, our model outperforms the other methods as evidenced in Table~\ref{tab:main_table}. 
We include examples of reconstructed images in the Supplement.

\footnotetext{Due to the lack of shared code and details in the AttSeg and SimGlim works, we are not able to estimate the required GFLOPs.}

\subsection{Ablations and analysis}
Firstly, we examine the analysis of architectural choices and their impact on the model's performance concerning classification accuracy and computational costs. Subsequently, we explore how randomized choice of $\kappa$ during training enables $\tore$ effective and flexible accuracy-efficiency tradeoff. Following this, we assess the model's effectiveness throughout the exploration process. Then, we investigate how the TORE forward pass influences exploration in AVE. Finally, we compare the efficiency and accuracy of $\tore$ and EfficientFormer~\cite{li2022efficientformervisiontransformersmobilenet}, and we provide an analysis of why EfficientFormer cannot be easily incorporated into the AVE setup.

\vspace{-1em}
\paragraph{What is the impact of particular design choices on the TORE's performance?}
Table~\ref{tab:ablation} presents the impact of various components of the TORE method on its performance. The results underscore the necessity of all components for achieving the best-performing model. Training with the random glimpse selection policy enhances accuracy by 10\% while the utilization of a lightweight decoder reduces the FLOPs requirement for AVE by threefold and additionally improves accuracy by 4\%.

\begin{table*}[t]

    \centering
    \begin{tabular}{ccccccc}
        \toprule
        \multirow{2}{*}{TORE FP} & Random & Lightweight & \multicolumn{2}{c}{SUN360}& \multicolumn{2}{c}{CIFAR100} \\ 
        &GSP&Decoder& Acc. $\uparrow$ & GFLOPs $\downarrow$ & Acc. $\uparrow$ & GFLOPs $\downarrow$\\ 
        \midrule
        \xmark & \cmark & \cmark & \textbf{77.1} & \textbf{13.4} & \textbf{80.2} & \textbf{28.6} \\
        \xmark & \xmark & \cmark & 70.3 & \textbf{13.4} & 73.8 & \textbf{28.6} \\
        \xmark & \cmark & \xmark & 72.2 & 38.8 & 76.1 & 87.0 \\ \midrule
        
        \cmark & \cmark & \cmark & \textbf{74.1} & \textbf{7.0} & \textbf{73.5} & \textbf{15.9}\\
        \cmark & \xmark & \cmark & 68.8 & \textbf{7.0} & 65.6 & \textbf{15.9} \\
        \cmark & \cmark & \xmark & 68.2 & 32.4 & 69.8 & 73.8 \\ \bottomrule
    \end{tabular}
        \caption{Ablation analysis of TORE. The ablation reveals that all TORE components are necessary to achieve the best-performing model. Note that the shortcuts correspond to TORE Forward Pass (TORE FP) and Random Glimpse Selection Policy (Random GSP).}

    \label{tab:ablation}
    \vspace{-1em}
\end{table*}

Note that, we examine the model's prediction accuracy for $\kappa=0$ in the TORE forward pass, where the model operates similarly to the original ViT, and for $\kappa>0$, where the model operates within the \iterator-\aggregator framework. In the latter case, we set $\kappa=8$ for SUN360 and $\kappa = 7$ for CIFAR100, as they achieve comparable classification performance to AME as presented in Table~\ref{tab:main_table}.

 \begin{figure}[t]
     \begin{center}
 \includegraphics[width=\linewidth]{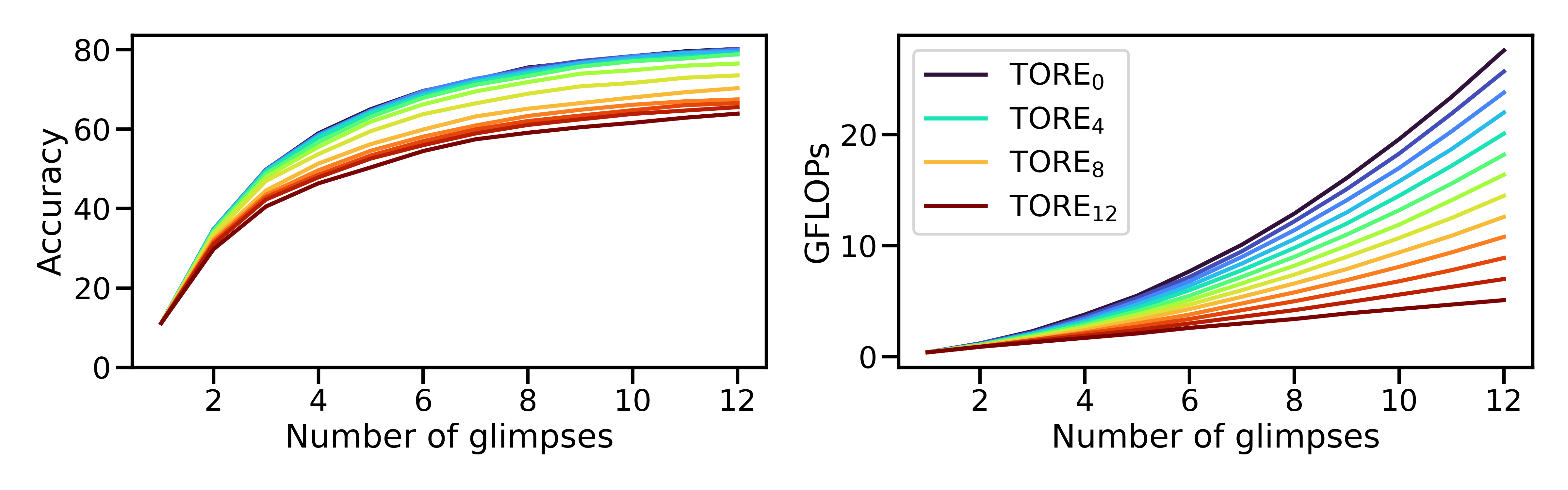}
   \end{center}
    \vspace{-2em}
     \caption{TORE accuracy (left) and resource utilization (right) with respect to the number of exploration steps on CIFAR100. The accuracy consistently improves with the increasing number of exploration steps and decreasing value of $\kappa$ ranging from 60\% to 80\% at the 12th step. Interestingly, minor drops in accuracy (e.g., $\text{TORE}_4$) correspond to a significant 20\% reduction in computational cost.}
     \label{fig:seq_acc_k}
     \vspace{-1em}
 \end{figure}

\vspace{-1em}
\paragraph{How does random $\kappa$ sampling during training affect model performance?}
The results in Table~\ref{tab:fixed_vs_random} reveal that the training schema with randomized $\kappa$ values enables the model to robustly conserve classification performance for greater values of $\kappa$ with only marginal performance decrease in the base case ($\kappa = 0$). Notably, the single model trained with random $\kappa$ sampling outperforms dedicated models trained for specific values of $\kappa = \kappa_{eval}$.

\vspace{-1em}
\paragraph{How does the model perform throughout a whole exploration process?} 

We record model predictions at each timestep of exploration and report its accuracy. We conduct this experiment for each value of $\kappa$ and plot the gathered results in Figure \ref{fig:seq_acc_k}. The accuracy consistently improves with the increasing number of exploration steps and decreasing value of $\kappa$, and ranges from 60\% to 80\% at the 12th step on the CIFAR100 dataset.

\begin{table}[t]
    \centering
    \begin{tabular}{c|ccccc}
         $\kappa$ sampling & \multicolumn{5}{c}{Evaluation $\kappa_{eval}$} \\ 
         in  training & 0 & 2 & 4 & 6 & 8 \\ \hline
        fixed $(\kappa=\kappa_{eval})$ & \textbf{77.4} & 73.6 & 71.6 & 73.8 & 71.7 \\
         random & 77.1 & \textbf{76.2} & \textbf{74.1} & \textbf{74.1} & \textbf{74.1}
    \end{tabular}
    
    \caption{Comparison of $\tore$ trained with fixed $\kappa$ or random $\kappa$ sampling on SUN360. The single, flexible model trained with random $\kappa$ sampling is superior to specialized models trained for specific values of $\kappa$.}
    \label{tab:fixed_vs_random}
\vspace{-1em}
\end{table}

\vspace{-1em}
\paragraph{What is the impact of TORE forward pass on exploration quality and its predictive power?}

In Active Visual Exploration, the model performance relies on two components: the predictive power of the model and the selected glimpses during exploration. We analyze the impact of those two components by analyzing the model accuracy on sets of patches gathered by different exploration processes.

\begin{figure}[t]
  \begin{center}
 \includegraphics[width=\linewidth]{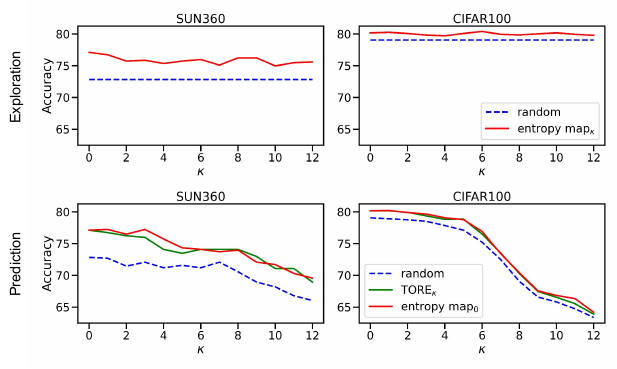}
  \end{center}
  
    \vspace{-2em}
     \caption{The impact of TORE forward pass on exploration and prediction quality. We show that using TORE forward pass does not impede exploration for any value of $\kappa$ and results in only a marginal decrease in prediction accuracy for $\kappa < 6$.
     At the top, we measure TORE$_{0}$ accuracy on glimpses gathered with the random policy (blue dashed line) and with the policy based on the entropy maps produced by TORE$_{\kappa}$ (red solid line) for different values of $\kappa$. At the bottom, we measure TORE$_{\kappa}$ accuracy for different $\kappa$ values on glimpses gathered with the random policy (blue dashed line) and the policies based on the entropy maps produced by TORE$_{\kappa}$ (green solid line)  and by TORE$_{0}$ (red solid line).
     }
     \label{fig:impact_on_explo}
     \vspace{-1em}
 \end{figure}

In the first case, we measure model accuracy with the original ViT forward pass ($\kappa=0$) on glimpses gathered by a model operating in \iterator-\aggregator framework with different $\kappa$ values, see upper row of Figure~\ref{fig:impact_on_explo}. We observe that glimpses selected using entropy maps are of similar quality for all $\kappa$ and better than the randomly chosen ones.

In the second case, we measure the accuracy of the model operating in \iterator-\aggregator framework with different values of $\kappa$ on glimpses gathered by a model using the original ViT forward pass ($\kappa = 0$), see the bottom row in Figure \ref{fig:impact_on_explo}.
We observe that for all values of $\kappa$, the model achieves higher accuracy on glimpses selected based on entropy maps than the ones chosen at random. For comparison, we plot the results for TORE$_{\kappa}$, which predicts and gathers glimpses in \iterator-\aggregator framework using the same values of $\kappa$. Note that the accuracy of TORE is almost the same as the accuracy of a model predicting on glimpses gathered with $\kappa=0$.

Results in both cases show that the model using TORE forward pass can utilize the entropy maps of the reconstruction decoder effectively. Additionally, the value of $\kappa$ does not significantly impact the quality of chosen patches. We conclude that the TORE forward pass is well suited for Active Visual Exploration with policies based on the properties of the attention map.

\begin{table}[t]
    \small
    \centering
    \begin{tabular}{c|cccc|c}
         Method & SUN$^*$ & CIFAR  & Flowers & Food & GFLOPS\\ \hline
         EF & 74.1 & 79.7 & 64.1 & \textbf{81.4} & 31.1 \\
         $\tore_0$ & \textbf{78.6} & \textbf{80.2} & \textbf{86.3} & 80.4 & 29.1 \\
         $\tore_6$ & 75.7 & 76.5 & 84.4 & 78.4 & \textbf{16.4}\\
         
    \end{tabular}
    \caption{Results of $\tore$, and EfficientFormer (EF) trained for AVE task on $4$ datasets. $\tore_0$ outperforms EF with respect to both accuracy and efficiency on 3 out of 4 evaluated datasets. Note that, we used the SUN360 dataset with images resized to $224\times 224$ resolution because EF is designed to process square images.}
    \label{tab:comp_efficient}
    \vspace{-1em}
\end{table}

\vspace{-1em}
\paragraph{Is $\tore$ more effective than optimized for computational efficiency ViT architectures such as EfficientFormer?} 
Reducing computational costs for Vision Transformers (ViTs) can be achieved by designing efficient ViT architectures~\cite{cai2022efficientvit, li2022efficientformervisiontransformersmobilenet}. However, AVE is a specific setup that processes incomplete and unstructured input, requiring data imputation for efficient ViT variants due to their altered attention mechanisms. Traditional ViT architectures, on the other hand, are advantageous as they can process only the available image patches at any given time without requiring data imputation.

To demonstrate that $\tore$ is more efficient than computationally effective ViT versions, we trained the EfficientFormer-V2-L (EF) in the same manner as $\tore$. This means that we use a small reconstruction decoder and a random glimpse selection policy during training, with exploration based on the entropy of attention maps (AME). The only difference is that EF processes the same number of tokens in each iteration. Those tokens that are not discovered through the exploration are imputed as black. This ensures that the input is of a fixed size for computation and allows a fair comparison.

The results in Table~\ref{tab:comp_efficient} indicate that TORE outperforms EF on 3 out of 4 evaluated datasets and requires fewer FLOPS. Even with $\kappa=0$, TORE is more efficient because it uses only visible tokens instead of a fully-sized imputed image at each exploration step.
\vspace{-1em}
\section{Conclusions and future works}

In this work, we introduce the TOken REcycling (\method) approach for enhancing efficiency in Active Visual Exploration tasks. It involves an efficient sequential inference that divides the ViT backbone into two components: \iterator and \aggregator. Through the concept of splitting the inference into two paths, we can fully utilize the potential of pretrained models. Additionally, we propose the use of a lightweight decoder in AVE, demonstrating that reduced reconstruction performance does not necessarily compromise the model's accuracy. Finally, we propose a training schema aimed at improving the model's downstream performance. As a result, \method significantly reduces the computational workload while maintaining or even enhancing accuracy in Active Visual Exploration tasks.

In future research, we aim to explore further reductions in computations, such as modifying the \aggregator by integrating an attention-pooling mechanism. Additionally, we plan to refine the proposed framework by incorporating early exits to further alleviate the computational burden.

\vspace{-1em}
\paragraph{Limitations.}
The primary limitation of the study lies in the fixed nature of the image divisions and masks used in the experiments, constrained by the size of the image patches in the ViT model. However, in future work, we will explore the impact of more random patch sizes on the model to better understand the model's behavior.

\vspace{-1em}
\paragraph{Impact.}
This work impacts the fields of embodied AI, robot vision, and efficient machine learning. It shows the potential of a straightforward yet powerful modification in the forward pass and training of ViTs, significantly reducing the computational load of large models and enabling efficient computations on devices such as drones. 

\vspace{-1em}
\section*{Acknowledgements}
The work of Jan Olszewski and Mateusz Pach was funded by the National Science Centre (Poland) grant no. 2022/47/B/ST6/03397. Dawid Rymarczyk was supported by the National Science Centre (Poland), grant no. 2022/45/N/ST6/04147. The work of Bartosz Zieliński was supported by the National Science Centre (Poland), grant no. 2023/50/E/ST6/00469.

We gratefully acknowledge Polish high-performance computing infrastructure PLGrid (HPC Centers: ACK Cyfronet AGH) for providing computer facilities and support within
computational grant no. PLG/2023/016555.

Some experiments were performed on servers purchased with funds from a grant from the Priority Research Area (Artificial Intelligence Computing Center Core Facility) under the Strategic Programme Excellence Initiative at Jagiellonian University.
\bibliographystyle{splncs04}
\bibliography{main}

\clearpage

\onecolumn
\section*{TORE: Token Recycling in Vision Transformers for Efficient Active Visual Exploration -- Supplement}

\begin{figure*}
    \centering
    \includegraphics[width=\textwidth]{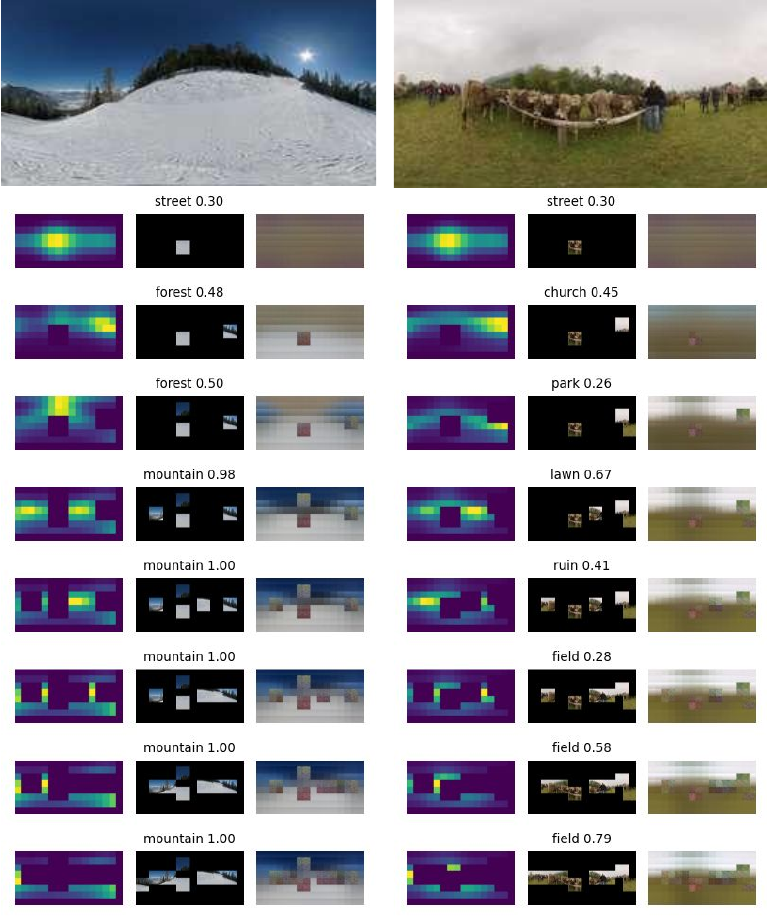}
    \caption{Visualization of active visual exploration performed by TORE. For each image, in the first column, we present entropy maps used by the selection policy, in the second column, the model successive fields of view, and finally in the last column we show the image reconstructions at each timestep. Note that we train TORE using masked reconstruction loss, therefore the model is not required to produce meaningful reconstructions for visible patches.}
    \label{fig:supp}
\end{figure*}

We provide examples of full exploration trajectories performed by our method as illustrated in \ref{fig:supp}. Although image reconstruction is of poor quality, it is sufficient to guide the exploration process and results in the high accuracy of our method on the AVE task.

Additionally, we provide the source code to reproduce the results presented in the paper.

\end{document}